# SOME THEORETICAL RESULTS ON DISCRETE CONTOUR TREES

*Yuqing Song*

Tianjin University of Technology and Education
Tianjin 300222, China, yqsong7@hotmail.com

**ABSTRACT**

Contour trees have been developed to visualize or encode scalar data in imaging technologies and scientific simulations. Contours are defined on a continuous scalar field. For discrete data, a continuous function is first interpolated, where contours are then defined. In this paper we define a discrete contour tree, called the iso-tree, on a scalar graph, and discuss its properties. We show that the iso-tree model works for data of all dimensions, and develop an axiomatic system formalizing the discrete contour structures. We also report an isomorphism between iso-trees and augmented contour trees, showing that contour tree algorithms can be used to compute discrete contour trees, and vice versa.

## 1. INTRODUCTION

In image processing and many other fields, contour trees have been used extensively for analysis, visualization, and simulation [1]-[5]. For a continuous function $f$, a contour is defined to be a component of a level set $\{x \mid f(x) = v\}$ for some value $v$. For discrete data, a continuous function is interpolated, where contours are defined. To ensure that the critical points are isolated, a perturbation is applied to the input values to make them unique. A contour tree is a graph consisting of contours passing through the critical points, and it can be augmented with the contours passing through the regular points.

Contour trees are defined on continuous functions, but the input to the computing algorithms is indeed a scalar graph consisting a point set, an adjacency set, and a scalar function on the points. Two questions naturally arise: whether we can define contour trees on graphs, and if yes, on what kind of graphs the contours of a function are always nested to make a tree.

Monasse et al. proposed the tree of shapes (ToS) on semicontinuous functions: a component (of either an upper- or lower- level set) with its holes filled is called a shape, which can be determined by its boundary (called a level line/surface); all shapes are nested, making a rooted tree, called the tree of shapes [6][7], which is a fusion of the dual component trees [8][9], and has been applied widely [10]-[14]. The boundary of a shape in the ToS is a semicontinuous equivalent of a contour.

Our previous work [15] introduced on the hexagonal grid the concept of monotonic line, which is a boundary where the input 2D image assumes higher/lower values on the pixels adjacent to the boundary from inside than those from outside. A monotonic line of higher/lower values on immediate inside is a discrete equivalent of a positive/negative contour line in [1]. All monotonic lines are nested, forming a rooted tree, called the monotonic tree, which was used in image analysis and retrieval [16][17]. We later adopted the level line term, renamed monotonic trees to level line trees, and proposed topdown algorithms to compute the trees on the hexagonal grid [18][19] and the square grid [20].

This paper extends the monotonic/level line tree definition from the hexagonal and square grids to mono-connected graphs, and defines an iso-tree as a free tree consisting of level cuts as the edges of the tree. Each level cut is a discrete contour, which is called a level line/surface on a 2D/3D grid [6][7][18][20]. All level cuts divide the domain into a set of iso-zones, which are the vertices of the iso-tree. The novelty of the paper is summarized as follows.

- We define the mono-connectivity of graphs, and show that a simplicial mesh is mono-connected if it is a triangulation of a manifold homeomorphic to a closed unit ball of any dimensions, suggesting that the iso-tree model works for data of all dimensions.
- We define iso-trees on graphs, and show that the mono-connectivity of a graph ensures a tree structure of the contours. Existing works define contours on geometric/topological spaces or grids, but not directly on a general graph.
- We report an isomorphism between iso-trees and augmented contour trees, assuring that contour tree algorithms can be used to compute discrete contour trees, and vice versa.

- We develop an axiomatic system to formalize the level cut structures. The iso-tree definition and the axiomatic formalization on graphs offer a base and inspirations for investigating discrete problems in a discrete way.

The paper is organized as follows. Section 2 is dedicated to basic notions and manipulations of mono-connected graphs. Section 3 elaborates the level cuts and the iso-tree of a scalar mono-connected graph. Section 4 makes an axiomatic formalization to find the basic properties necessary and sufficient to model iso-trees. Section 5 reports an isomorphism between the iso-tree and the augmented contour tree. Last we conclude in Section 6. Due to space constrains, all proofs are omitted and will be provided in a forthcoming document.

## 2. MONO-CONNECTED GRAPH

This section defines the mono-connectivity of graphs and discusses its properties.

On a graph $<\Omega, \mathfrak{E}>$, except the following term aliases, we adopt common terminologies of graph theory in our formulation.
(1) Each vertex in $\Omega$ is called an *input site*, or simply a *site*.
(2) Each edge in $\mathfrak{E} \subseteq \Omega \times \Omega$ is called a *surfel* (short for *surface element*) [21].

We impose no geometrical restriction on the graph. The following geometrical interpretation is introduced to help understanding, and itself is not part of the theory. An input site can be perceived as the Voronoi neighborhood of the site (for a Delaunay triangulation) or a voxel (short for volume element, for a general graph). A surfel noted by a pair $(p, q)$ can be interpreted as a face shared by the voxels of the adjacent sites $p$ and $q$. In this paper when depicting a graph in a figure, we use this interpretation. See Figure 1.

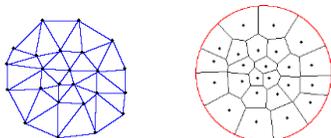

Figure 1: (**a-left**) A graph. (**b-right**) The geometrical interpretation. The right diagram is the dual Voronoi Tessellation of the left Delaunay Triangulation.

A subset $X$ of $\Omega$ is called a *region*; its *complement*, $\Omega - X$, is noted by $X^c$. If $X$ is connected, a component of $X^c$ is called a *cavity* of $X$. The *immediate interior* (*II*) of $X$ is defined as $II(X) = \{p \in X \mid \exists (p, q) \in \mathfrak{E} \text{ such that } q \in X^c\}$, which is the set of the boundary sites inside. The set of the boundary sites outside can be expressed as $II(X^c)$. A *Jordan cut* (or simply a $\mathcal{J}$-*cut*) $\alpha$ is a bipartition of $\Omega$, represented by a pair $(X, X^c)$, such that both $X$ and $X^c$ are non-empty and connected regions. The *boundary* of $\alpha$ is defined as the $\partial \alpha = \{(p, q) \in \mathfrak{E} \mid p \in X \text{ and } q \in X^c\}$. A graph is called *mono-connected*, if it is connected, and for any $\mathcal{J}$-cut $(X, X^c)$ of the graph, both $II(X)$ and $II(X^c)$ are connected. See Figure 2.

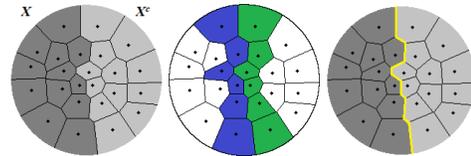

Figure 2: (**left**) a $\mathcal{J}$-cut (in deep/shallow grays) on a mono-connected graph. (**mid**) $II(X)$ in blue and $II(X^c)$ in green. (**right**) The boundary of the $\mathcal{J}$-cut in yellow.

The mono-connectivity (short for mono-channel connectivity) is a discrete equivalent of the simple-connectivity on topological spaces. A simply-connected space is a space where there is only one channel to connect two points, and any Jordan cut has a connected boundary. See Figure 3.

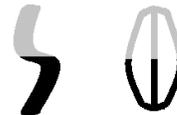

Figure 3: A Jordan cut in black/gray on (**left**) a simply-connected space, where the boundary of the cut is connected, (**right**) a not simply-connected space, where the boundary of the cut is not connected.

In the rest of the paper, we let $\mathfrak{S} = <\Omega, \mathfrak{E}>$ **be a fixed mono-connected graph**, on which we define the iso-tree and all related concepts.

In the graph $\mathfrak{S}$, surfels are oriented: $(p, q)$ and $(q, p)$ are inverse to each other, i.e., $(p, q)^{-1} = (q, p)$. The *inverse* of a surfel set $S$ is $S^{-1} = \{e^{-1} \mid e \in S\}$.

Let $\alpha = (X, X^c)$ be a $\mathcal{J}$-cut of $\mathfrak{S}$. The $\mathcal{J}$-cut is an oriented surface dividing the domain into two regions $X$ and $X^c$. We name $X$ (resp. $X^c$) the *low-* (resp. *up-*) *teriors*[1] of the $\mathcal{J}$-cut. A $\mathcal{J}$-cut $(Y, Y^c)$ is called *on the low-* (resp. *up-*) *side* of $\alpha$ if $Y \subsetneq X$ or $Y^c \subsetneq X$ (resp. $Y \subsetneq X^c$ or $Y^c \subsetneq X^c$). In general, an item (e.g., a region or a site) is called on *the low-* (resp. *up-*) *side* of $\alpha$ if the item is in $X$ (resp. $X^c$). Two items are called *separated* by $\alpha$ if they are in $X$ and $X^c$, respectively. The following theorem shows that the iso-tree model works for data of all dimensions.

**Theorem 1:** A simplicial mesh is mono-connected if the mesh is a triangulation of a manifold homeomorphic to a closed unit ball $\{x \in \mathbb{R}^d \mid \|x\| \leq 1\}$ for any $d > 1$.

## 3. LEVEL CUTS AND ISO-TREE

The graph $\mathfrak{S} = <\Omega, \mathfrak{E}>$ equipped with a scalar function $f$ (a real-valued function on $\Omega$) is called a *scalar graph*, represented as $<\Omega, \mathfrak{E}, f>$ and noted by $\mathfrak{S}_f$. See Figure 4(a). In this section we define and discuss the level cuts and the iso-tree of $\mathfrak{S}_f$.

---

[1] The names are just for distinguishing the two sides of the cut. Any meaningful names meet this need, such as "in/ex-teriors". To be consistent in $\mathcal{J}$-cuts and level cuts (to be defined in Section 3), "low/up-teriors" are used.

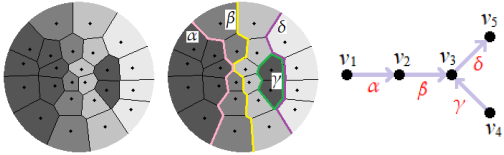

Figure 4: (**a-left**) A scalar graph, where the scalar values are depicted by different grays. (**b-mid**) The 4 level cuts $\alpha$, $\beta$, $\gamma$, and $\delta$. (**c-right**) The iso-tree, where the vertices are noted $v_1$, $v_2$,…, and $v_5$, respectively.

A *level cut* (or simply an $\mathcal{L}$-*cut*) of $\mathfrak{S}_f$ is a $\mathcal{J}$-cut $(X, X^c)$ such that max $f(\Pi(X))$ < min $f(\Pi(X^c))$. See Figure 4(b). The set of the level cuts of $\mathfrak{S}_f$ is noted by $\mathcal{L}_f$.

Theorem 2 lists two properties of $\mathcal{L}$-cuts. The first property affirms that $\mathcal{L}$-cuts are nested. Put another way: they do not cross each other, i.e., one of the 4 intersections is empty: $X \cap Y$, $X \cap Y^c$, $X^c \cap Y$, and $X^c \cap Y^c$. The second property prevents $\mathcal{L}$-cuts from being inversely tangent. Two $\mathcal{J}$-cuts, $\alpha$ and $\beta$, are called *tangent* if $(\partial \alpha) \cap (\partial \beta) \neq \varnothing$; they are called *inversely tangent* if $(\partial \alpha) \cap (\partial \beta)^{-1} \neq \varnothing$. In Figure 4, the $\mathcal{L}$-cuts $\alpha$ and $\beta$ are tangent; so are $\gamma$ and $\delta$.

**Theorem 2:** Any $\mathcal{L}$-cuts $(X, X^c)$ and $(Y, Y^c)$ in $\mathcal{L}_f$ have the following properties,
- *Nesting Property*: $X \subseteq Y$, $X \subseteq Y^c$, $X^c \subseteq Y$, or $X^c \subseteq Y^c$,
- *Tangent Property*: $\partial(X, X^c) \cap (\partial(Y, Y^c))^{-1} = \varnothing$.

Based on the properties discussed, we derive the iso-tree in the following theorem.

**Theorem 3:** All $\mathcal{L}$-cuts in $\mathcal{L}_f$ make a free tree where each $\mathcal{L}$-cut $\alpha$ in $\mathcal{L}_f$ is an edge of the tree connecting two subtrees $\tau_1$ and $\tau_2$ such that $\tau_1/\tau_2$ consists of the $\mathcal{L}$-cuts on the low/up-side of $\alpha$.

The free tree described in Theorem 3 is called the *iso-tree* of the scalar graph $\mathfrak{S}_f$. See Figure 4(c). The iso-tree is indeed a directed tree and we let an edge directed from its low side to its up side. The two vertices connected by an edge are called the *low* and *up vertices* of the edge, respectively.

A vertex of the iso-tree is called an $\mathcal{L}$-*vertex* of $\mathcal{L}_f$. An $\mathcal{L}$-vertex $v$ has a set $C$ of incident edges, each of which is an $\mathcal{L}$-cut. If $v$ is the low (resp. up) vertex of an incident $\mathcal{L}$-cut $\alpha \in C$, then the low- (resp. up-) terior of $\alpha$ is called the $v$-including terior of $\alpha$, denoted by $inTerior(\alpha, v)$. The set $\cap\{inTerior(\alpha, v) \mid \alpha \in C\}$ is a region bounded by the incident $\mathcal{L}$-cuts of $v$, called the *zone* of $v$. For the example in Figure 4(c), [2] $Zone(v_3) = inTerior(\beta, v_3) \cap inTerior(\gamma, v_3) \cap inTerior(\delta, v_3) = upTerior(\beta) \cap upTerior(\gamma) \cap lowTerior(\delta)$. See Figure 5. Based on Theorem 3, we can also prove the following properties of the zones.

---
[2] In this paper, unless otherwise noted, a property of an item $t$ is denoted by ***propertyName***($t$), such as the "$Zone(v_3)$" here and the "*valueGap*($\alpha$)" in Section 4.1, with the first letter of the last word capitalized.

**Corollary 1**: On $\mathcal{L}_f$,
(1) The zones of the $\mathcal{L}$-vertices make a partition of the domain $\Omega$ (see Figure 5).
(2) The zone of any $\mathcal{L}$-vertex is not empty, and can be disconnected.
(3) The function $f$ is constant on the zone of any $\mathcal{L}$-vertex.

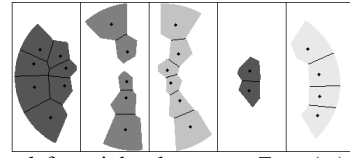

Figure 5: From left to right, the zones, *Zone* ($v_1$), *Zone* ($v_2$), …, *Zone* ($v_5$), of the $\mathcal{L}$-vertices in Figure 4(c).

The zone of an $\mathcal{L}$-vertex of $\mathfrak{S}_f$, is called an *iso-zone* (a zone of a constant value). The constant value of $f$ on an iso-zone is called the *value* of the iso-zone. By Corollary 1, an iso-zone cannot be empty, and there is a one-to-one correspondence between the $\mathcal{L}$-vertices and the iso-zones, and we will use iso-zone as an alias of $\mathcal{L}$-vertex. The two iso-zones connected by an $\mathcal{L}$-cut $\alpha$ are respectively called the *low and up zones* of $\alpha$ (whether low/up depends on which side they belong to). The value of the low (resp. up) zone is called the *low* (resp. *up*) *value* of $\alpha$. The difference between the low and up values is called the *value gap* of $\alpha$.

## 4. AXIOMATIC FORMALIZATION

An axiomatic system is a system which possesses an explicitly stated set of axioms from which theorems can be derived [22]. Our axiomatic formalization aims to find the basic properties necessary and sufficient to model iso-trees so that all other properties can be deduced from the basic ones.

We start with a concise representation of iso-trees and will show in this section that the nesting and tangent properties listed in Theorem 2 are the basic properties to model iso-trees.

### 4.1. Iso-tree Representation and Input Reconstruction

On the graph $\mathfrak{S}$, we let $r$ be a fixed reference site, and consider the function set $\{ f : \Omega \to \mathbb{R} \mid f(r) = 0 \}$ [3], noted by $\mathfrak{F}$. The iso-trees of all $f \in \mathfrak{F}$ make a set, noted by $\mathcal{T}$. Each iso-tree in $\mathcal{T}$ consists of iso-zones and $\mathcal{L}$-cuts, where the iso-zones can be derived from the $\mathcal{L}$-cuts as shown by Theorem 3. A concise representation of a $\tau \in \mathcal{T}$ is a set of $\mathcal{L}$-cuts, where each $\mathcal{L}$-cut is a $\mathcal{J}$-cut associated with a value gap. To prove that this representation is complete, we show how to recover the input function from the representation. We first derive an expression to recover the value of an iso-zone. For the example in Figure 4, if the reference site is in *Zone* ($v_1$), then $Value(Zone(v_4)) = valueGap(\alpha) + valueGap(\beta) - valueGap(\gamma)$. Formally, let

- $\tau \in \mathcal{T}$ be arbitrary,
- $z$ be an arbitrary iso-zone of $\tau$,

---
[3] The value of the reference site can be an arbitrary real number. W.l.o.g., we let it be 0.

- $z_r$ be the iso-zone of $\tau$ containing $r$,
- $\alpha_1, \alpha_2, \ldots, \alpha_m$ be the $\mathcal{L}$-cuts in the directed path from $z_r$ to $z$,
- *IsBackward* ($i$) be 1 (true) / 0 (false) if the direction of $\alpha_i$ is backward/forward in the path.

Then the recovered value of the iso-zone $z$ is expressed as
$$\sum_{i=1 \text{ to } m} (-1)^{IsBackward(i)} valueGap(\alpha_i).$$
We define the recovered function $h$ as: for any iso-zone $z$ of $\tau$ and any $p \in z$, $h(p)$ is equal to the recovered value of $z$.

With the properties discussed, we can easily establish that the recovered function is equal to the input function. Formally, the *iso-tree transform*, denoted as *ITT*, is a mapping from $\mathcal{F}$ to $\mathcal{T}$ such that for each $h \in \mathcal{F}$, $ITT(h)$ is the iso-tree of $\mathfrak{S}_h$. The *reconstruction transform*, denoted as *RT*, is a mapping from $\mathcal{T}$ to $\mathcal{F}$ such that for any iso-tree $\tau \in \mathcal{T}$, $RT(\tau)$ is the function recovered from $\tau$. The correctness of the recovering process is expressed by the following obvious theorem.

**Theorem 4:** *ITT*: $\mathcal{F} \to \mathcal{T}$ and *RT*: $\mathcal{T} \to \mathcal{F}$ are inverse transforms to each other.

### 4.2. Regular $\mathcal{J}$-Division

We further discuss the iso-tree representation to find the axioms for our axiomatic system. A set of $\mathcal{J}$-cuts on $\mathfrak{S}$ is called a *Jordan division*, or simply $\mathcal{J}$-*division*, of $\mathfrak{S}$. A *valued $\mathcal{J}$-division* of $\mathfrak{S}$ is a pair $<D, V_G>$ where $D$ is a $\mathcal{J}$-division of $\mathfrak{S}$, and $V_G$ is a positive function on $D$, assigning each $\mathcal{J}$-cut a value gap. Obviously an iso-tree is a valued $\mathcal{J}$-division, but a valued $\mathcal{J}$-division may not induce an iso-tree. To qualify for an iso-tree, a valued $\mathcal{J}$-division needs to meet some requirements. Our investigation shows that the nesting and tangent properties listed in Theorem 2 constitute the qualifying criteria, and therefore make the axioms we are looking for:

A $\mathcal{J}$-division of $\mathfrak{S}$ is called *regular* if it satisfies the following axioms, for any $\mathcal{J}$-cuts $(X, X^c)$ and $(Y, Y^c)$ in it,
- *Nesting Axiom*: $X \subseteq Y$, $X \subseteq Y^c$, $X^c \subseteq Y$, or $X^c \subseteq Y^c$.
- *Tangent Axiom*: $\partial(X, X^c) \cap (\partial(Y, Y^c))^{-1} = \varnothing$.

Regular $\mathcal{J}$-division will hereafter be abbreviated as $\mathcal{J}_r$-division. We let $\mathcal{D}$ be the set of the valued $\mathcal{J}_r$-divisions of $\mathfrak{S}$.

As Theorem 2 states, iso-trees are valued $\mathcal{J}_r$-divisions, i.e., $\mathcal{T} \subseteq \mathcal{D}$, affirming the necessity of the nesting and tangent axioms for the axiomatic system. To show they are sufficient, it suffices to prove: valued $\mathcal{J}_r$-divisions are iso-trees i.e., $\mathcal{D} \subseteq \mathcal{T}$, as stated in Theorem 5.

**Theorem 5:** On the graph $\mathfrak{S}$, iso-trees are valued $\mathcal{J}_r$-divisions, and vice versa.

## 5. ISOMORPHISM TO AUGMENTED CONTOUR TREE

In this section we will equate the iso-tree with the augmented contour tree on a scalar triangulation of a $d$-manifold homeomorphic to a closed unit ball for any $d > 1$. We use the contour tree definition in [2], and let the input graph $\mathfrak{S}$ be such a triangulation with $\Omega$ the vertex set and $\mathfrak{E}$ the adjacency set.

As required by the Morse theory, the input values must be unique, and if not, a perturbation is applied to make them so. Let $f$ be a real-valued scalar function on $\Omega$, $h$ be a function unique at each site which is made by a perturbation of $f$, and denote the scalar graphs $<\mathfrak{S}, f>$ and $<\mathfrak{S}, h>$ respectively by $\mathfrak{S}_f$ and $\mathfrak{S}_h$. We show that contour tree algorithms can be used to compute discrete contour trees, and vice versa, by proving the following results.
- The iso-tree of $\mathfrak{S}_h$ is isomorphic to the augmented contour tree of $\mathfrak{S}_h$ (Theorem 6).
- The iso-tree of $\mathfrak{S}_f$ can be obtained by a tree reduction on the iso-tree of $\mathfrak{S}_h$: each iso-zone of $\mathfrak{S}_f$ corresponds to a connected sub-graph of the iso-tree of $\mathfrak{S}_h$; we reduce each of these sub-graphs into one node by (1) examining each $\mathcal{L}$-cut in $\mathfrak{S}_h$, and (2) if its low-zone and up-zone are of the same $f$ value, then merging the two zones (Theorem 7).

**Theorem 6:** If a scalar function $h$ on $\Omega$ has different values at all sites, then the augmented contour tree $\tau_c$ and the iso-tree $\tau_i$ of $\mathfrak{S}_h$ are isomorphic such that each contour node in $\tau_c$ passing through a site $p$ is mapped to the iso-zone $\{p\}$ in $\tau_i$.

**Theorem 7:** If a scalar function $h$ on $\Omega$ is made by a perturbation of a function $f$ such that $h$ assumes a unique value at each site, let $\tau_i = <Z_i, L_i>$ be the iso-tree of $\mathfrak{S}_i$ with $Z_i$ the iso-zone set and $L_i$ the $\mathcal{L}$-cut set, for $i = f$ or $h$. Then
(1) $L_f \subseteq L_h$.
(2) $L_h - L_f = \{\alpha \in L_h \mid f(lowZone^h(\alpha)) = f(upZone^h(\alpha))\}$.[4]
(3) For any iso-zone $u \in Z_f$, and any $p_1, p_2 \in u$, if $p_1 \neq p_2$, then the singleton sets $\{p_1\}$ and $\{p_2\}$ are iso-zones in $\tau_h$ which are connected by a path consisting of $\mathcal{L}$-cuts in $L_h - L_f$.
(4) $\forall \alpha \in L_h - L_f$, $\exists u \in Z_f$ s.t. $lowZone^h(\alpha) \cup upZone^h(\alpha) \subseteq u$.

## 6. CONCLUSION

This paper extends the monotonic/level line tree definition from the hexagonal and square grids to mono-connected graphs, and defines the iso-tree as a free tree consisting of level cuts as the edges. All level cuts divide the domain into a set of iso-zones, which are the vertices of the iso-tree. An axiomatic formalization was undertaken, which finds the basic properties necessary and sufficient to model iso-trees. An isomorphism between the iso-tree and the augmented contour tree was reported.

As further work, the iso-trees on mono-connected graphs will be extended to iso-graphs (graphs of iso-zones, a discrete version of Reeb graphs) on general graphs. The key problems in the extension include what kind of restrictions should be imposed on the input graphs to ensure desirable contour structure, and how to define a discrete contour on a graph not mono-connected.

---

[4] The superscript is added to indicate which tree, $\tau_h$ or $\tau_f$, the zones belong to.